\documentclass[journal]{IEEEtran}
\ifCLASSINFOpdf
  \usepackage[pdftex]{graphicx}
  \graphicspath{{images/}}
\else
\fi
\usepackage[cmex10]{amsmath}
\usepackage{algorithmic}
\ifCLASSOPTIONcompsoc
  \usepackage[caption=false,font=normalsize,labelfont=sf,textfont=sf]{subfig}
\else
  \usepackage[caption=false,font=footnotesize]{subfig}
\fi
\usepackage{url}

\usepackage{diagbox}

\allowdisplaybreaks

\newcommand{\tabincell}[2]{\begin{tabular}{@{}#1@{}}#2\end{tabular}}
\input{latex.macro}

\begin{document}
%
\title{Deep Metric Learning for Practical Person Re-Identification}
%
%
%

\author{Dong~Yi,
        Zhen~Lei,~\IEEEmembership{Member,~IEEE}
        Stan~Z.~Li,~\IEEEmembership{Fellow,~IEEE}
\thanks{Dong Yi is with the National Laboratory of Pattern Recognition,
Institute of Automation, Chinese Academy of Sciences, Beijing, China, 100190
e-mail:dyi@cbsr.ia.ac.cn}
\thanks{Manuscript received June 1, 2014; revised June 1, 2014.}}

%
%

\markboth{Journal of \LaTeX\ Class Files,~Vol.~11, No.~4, December~2012}%
{Shell \MakeLowercase{\textit{et al.}}: Bare Demo of IEEEtran.cls for Journals}
%



\maketitle

\begin{abstract}
Various hand-crafted features and metric learning methods prevail in the field of person re-identification. Compared to these methods, this paper proposes a more general way that can learn a similarity metric from image pixels directly. By using a ``siamese'' deep neural network, the proposed method can jointly learn the color feature, texture feature and metric in a unified framework. The network has a symmetry structure with two sub-networks which are connected by Cosine function. To deal with the big variations of person images, binomial deviance is used to evaluate the cost between similarities and labels, which is proved to be robust to outliers.

Compared to existing researches, a more practical setting is studied in the experiments that is training and test on different datasets (cross dataset person re-identification). Both in ``intra dataset'' and ``cross dataset'' settings, the superiorities of the proposed method are illustrated on VIPeR and PRID.
\end{abstract}

\begin{IEEEkeywords}
Person Re-Identification, Deep Metric Learning, Convolutional Network, Cross Dataset
\end{IEEEkeywords}

\ifCLASSOPTIONpeerreview
\begin{center} \bfseries EDICS Category: 3-BBND \end{center}
\fi
%
\IEEEpeerreviewmaketitle

\section{Introduction}

The task of person re-identification is to judge whether two person images belong to the same subject or not. In practical applications, the two images are usually captured by two cameras with disjoint views. The performance of person re-identification is closely related to many other applications, such as cross camera tracking, behaviour analysis, object retrieval and so on. The algorithms proposed in this field are also overlapped with other fields in pattern recognition. In recent years, the performance of person re-identification has increased continuously and will increase further.

The essence of person re-identification is very similar to biometric recognition problems, such as face recognition. The core of them is to find a good representation and a good metric to evaluate the similarities between samples. Compared to biometric problems, person re-identification is more challenging due to the low quality and high variety of person images. Person re-identification usually need to match the person images captured by surveillance cameras working in wide-angle mode. Therefore, the resolution of person images are very low (\eg around $48\times128$ pixels) and the lighting conditions are unstable too. Furthermore, the direction of cameras and the pose of persons are arbitrary. These factors cause the person images under surveillance scenarios have two distinctive properties: large variations in intra class, and ambiguities between inter classes. In summary, the challenges of person re-identification come from the following aspects: camera view change, pose variation, non-rigid deformation, unstable illumination and low resolution.

However, another challenge is less studied in existing work, which is ``cross dataset person re-identification''. In practical systems, we usually collect some large datasets first and train a model on them. Then the trained model is applied to other datasets or videos for person re-identification. We call the training datasets as source domain and test datasets as target domain. The source and target datasets are totally different, because they are usually captured by different cameras under different environments, \ie have different probability distribution. A practical person re-identification algorithm should has good generalization with respect to the dataset changes. Therefore, cross dataset person re-identification is an important rule to evaluate the performance of algorithms in practice.

Since the pixels of person images are unstable, effective representations are important and needed for person re-identification. To this end, existing methods borrow many sophisticated features from other fields, such as HSV histogram, Gabor, HOG and so on. Based on the features, direct matching or discriminative learning are then used to evaluate the similarity. Existing methods mainly focus on the second step that is how to learn a metric to discriminate the persons. Many good metric learning methods have been proposed in this context, such as KISSME~\cite{Kostinger-CVPR-2012}, RDC~\cite{Zheng-TPAMI-2013} and so on.

The majority of existing methods include two separate steps: feature extraction and metric learning. The features usually come from two separate sources: color and texture, some of which are designed by hand, some of which are learned, and they are finally connected or fused by simple strategies. On the contrary, this paper proposes a new method to combine the separate modules together, learning the color feature, texture feature and metric in a unified framework, which is called as ``Deep Metric Learning'' (DML).

The main idea of DML is inspired by a ``siamese'' neural network~\cite{Bromley-NIPS-1993}, which is originally proposed for signature verification. Given two person images $\bfx$ and $\bfy$, we want to use a siamese deep neural network to assess their similarity $s=DML(\bfx, \bfy)$. Being more specific than the original work~\cite{Bromley-NIPS-1993}, our DML first abstracts the siamese network to two sub-networks, a connection function and a cost function (see Figure~\ref{fig:arch}), and then carefully design the architecture for person images. In this way, DML can adapt well to person re-identification. Denoting the connection function as $S$, the similarity equation of DML can be written as $s=DML(\bfx, \bfy)=S(B_1(\bfx), B_2(\bfy))$, where $B_1$ and $B_2$ denote the two sub-networks of DML. Depending on specific applications, $B_1$ and $B_2$ need to share or not to share their parameters.

Compared with existing person re-identification methods, DML has the following advantages:
\begin{enumerate}
\item DML can learn a similarity metric from image pixels directly. All layers in DML are optimized by a common objective function, which are more effective than the hand-crafted features in traditional methods;
\item The multi-channel filters learned in DML can capture the color and texture information simultaneously, which are more reasonable than the simple fusion strategies in traditional methods, \eg feature concatenation, sum rule;
\item The structure of DML is flexible that can switch between view specific and general person re-identification tasks by whether sharing the parameters of sub-networks.
\end{enumerate}

DML is tested on two popular person re-identification datasets, VIPeR~\cite{Gray-PETS-2007} and PRID 2011~\cite{Hirzer-IA-2011}, using the common evaluation protocols. The results show that DML outperforms or on a par with the state-of-the-art methods, such as \cite{Li-CVPR-2013}, and \cite{Zhao-ICCV-2013}. To appeal the practical requirements and evaluate the generalization of DML, we also conduct more challenging cross dataset experiments, which are training on i-LIDS~\cite{Zheng-CVPR-2011}, CUHK Campus~\cite{Li-CVPR-2013} and testing on VIPeR, PRID. The results of the cross dataset experiments are significantly better than existing methods~\cite{Ma-ICCV-2013} under similar experimental settings. To our knowledge, this is the first work to conduct strict cross dataset experiment in the field of person re-identification. Finally, we fuse the features learned from multiple datasets to improve the performance on VIPeR and PRID further.


\section{Related Work}

This work uses deep learning to learn a metric for person re-identification. Related works in four aspects are reviewed in this section: feature representation, metric learning for person re-identification, siamese convolutional neural network and cross dataset related methods.

Early papers mainly focus on how to construct effective feature representation. From 2005, numerous features are used or proposed for person re-identification~\cite{Javed-CVPR-2005}. The features can be divided into two categories: color based and texture based features. The most popular features include HSV color histogram~\cite{Farenzena-CVPR-2010, Li-CVPR-2013}, LAB color histogram~\cite{Zhao-CVPR-2013}, SIFT~\cite{Zhao-CVPR-2013}, LBP histogram~\cite{Li-CVPR-2013}, Gabor features~\cite{Li-CVPR-2013} and their fusion. Among these features, color has the most contribution to the final results. The recent advance in this aspect is the color invariant signature~\cite{Kviatkovsky-TPAMI-2013}. By combing color histogram, covariance feature and segmentation information, \cite{Kviatkovsky-TPAMI-2013} achieved rank1=24\% on VIPeR.

On the other hand, \cite{Farenzena-CVPR-2010} has proved that using the silhouette and symmetry structure of person can improve the performance significantly, therefore the color and texture features are usually extracted in a predefined grid or finely localized parts. \cite{Xu-ICCV-2013} proposed a MCMC based method to do part localization and person re-identification simultaneously, which obtained rank1=23\% on VIPeR. The above work all prove that the performance can be improved significantly when know the geometric configuration of person explicitly or implicitly. Compared to part based method, salience based methods~\cite{Zhao-CVPR-2013, Zhao-ICCV-2013} proposed by Zhao \etal relaxed the spatial constraint further and could deal with larger pose variations. Similar to the history of face recognition~\cite{Zhang-PR-2009}, the future direction of feature representation must be based on precise body parts segmentation, person alignment and pose normalization.

Based on the extracted features, naive feature matching or unsupervised learning methods usually got moderate results, and  state-of-the-art results were achieved by supervised methods, such as Boosting~\cite{Gray-ECCV-2008}, Rank SVM~\cite{Prosser-BMVC-2010}, PLS~\cite{Prosser-BMVC-2010} and Metric learning~\cite{Zheng-TPAMI-2013, Kostinger-CVPR-2012, Li-CVPR-2013}. Among these methods, metric learning is the main stream due to its flexibility. Compared with standard distance measures, \eg $L_1$, $L_2$ norm, the learned metric is more discriminative for the task on hand and more robust to large variations of person images across view. Most papers used a holistic metric to evaluate the similarity of two samples, but \cite{Li-CVPR-2013} first divided the samples into several groups according to their pose and then learned metrics for each group. By using the pose information explicitly, \cite{Li-CVPR-2013} obtains the highest performance on VIPeR.
Recently, \cite{Liu-ICCV-2013} proposed a novel ``human in the loop'' style method, which illustrated that the performance of person re-identification can be improved drastically by human intervention. Although the results of this paper is hard to reproduce, it supplies a benchmark to reflect the performance of human (Rank1=71.08\% on VIPeR). Closing the gap to human performance is the target of researchers.

Early in 1993, a siamese neural network~\cite{Bromley-NIPS-1993} was proposed to evaluate the similarity of two signature samples. In the same year, a neural network~\cite{Baldi-NC-1993} with similar structure was proposed for fingerprint verification. Different from traditional neural networks, the siamese architecture is composed by two sub-networks sharing the same parameters. Each sub-network is a convolutional neural network. Then the siamese neural network was used for face verification~\cite{Chopra-CVPR-2005} by the same research group. The best property of siamese neural network is its unified and clear objective function. Guided by the objective function, the end-to-end neural network can learn a optimal metric towards the target automatically. The responsibility of the last layer of the siamese neural network is to evaluate the similarity of the output of two sub-networks, which can be in any form~\cite{Chopra-CVPR-2005}, such as $L_1$, $L_2$ norm and Cosine. Although good experimental results have been obtained in \cite{Bromley-NIPS-1993, Baldi-NC-1993} and \cite{Chopra-CVPR-2005}, their disadvantages are lacking implementation details and lacking comparison with other methods. This paper will design a siamese neural network for person image and apply it in the person re-identification problem. In this paper, the implementation details will be described and extensive comparisons will be reported.

In this young field, cross dataset problem has not attracted much attention. Majority researchers do their best to improve the performance within single dataset, \ie training on VIPeR and test on VIPeR too. Not long ago, \cite{Ma-ICCV-2013} started to concern this issue. In \cite{Ma-ICCV-2013}, the authors proposed a transfer Rank SVM (DTRSVM) to adapt a model trained on the source domain (i-LIDS or PRID) to target domain (VIPeR). All image pairs in the source domain and the negative image pairs in the target domain were used for training. Different from DTRSVM, in our cross dataset experiments, the proposed network is trained in the source domain only and its performance is tested in the target domain.

\section{Deep Metric Learning}

Under the joint influence of resolution, illumination and pose changes, the ideal metric for person re-identification may be highly nonlinear. Deep learning is exact one of the most effective tools to learn the nonlinear metric function. This section introduces the architecture, parameters, cost function and implementation details of the proposed convolutional network for deep metric learning.

\subsection{Architecture}

For most of pattern recognition problems, neural network works in a standalone mode. The input of neural network is a sample and the output is a predicted label. This mode works well for handwritten digit recognition, object recognition and other classification problems when the labels of the training set are the same as the test set. For person re-identification problem, the subjects in the training set are generally different from those in the test set, therefore the ``sample $\rightarrow$ label'' style neural network cannot apply to it. To deal with this problem, we construct a siamese neural network, which includes two sub-networks working in a ``sample pair $\rightarrow$ label'' mode.

\begin{figure}[!t]
\centering
\includegraphics[width=0.5\textwidth]{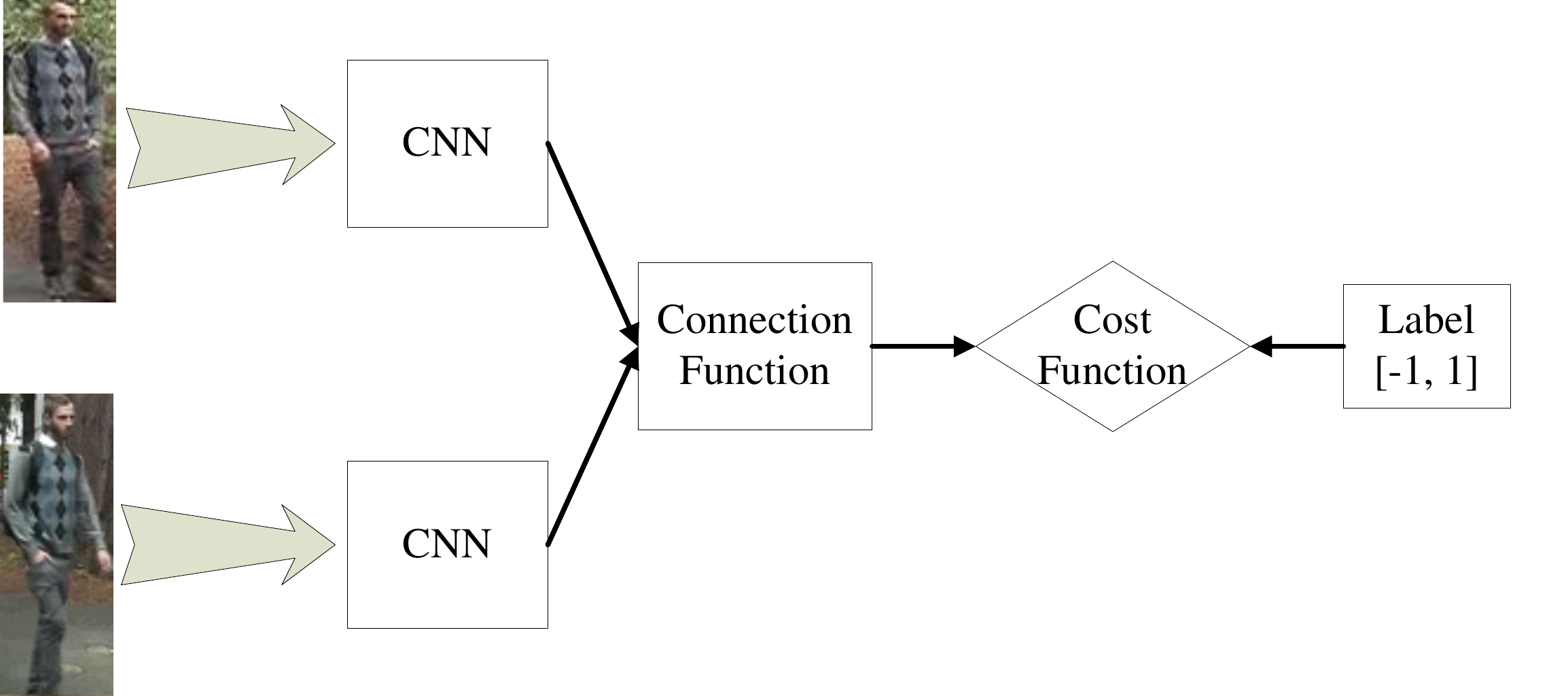}
\caption{The structure of the siamese convolutional neural network (SCNN), which is composed by three components: CNN, connection function and cost function.}
\label{fig:arch}
\end{figure}

The flowchart of our method is shown in Figure~\ref{fig:arch}. Given two person images, they are sent to siamese convolutional neural network (SCNN). For two images $\bfx$ and $\bfy$, SCNN can predict a label $l = \pm 1$ to denote whether the image pair comes from the same subject or not. Because many applications need rank the images in the gallery based on their similarities to a probe image, our SCNN outputs a similarity score instead. The structure of the SCNN is shown in Figure~\ref{fig:arch}, which is composed by two convolutional neural networks (CNN). And the two CNNs are connected by a connection function.

Existing siamese neural networks have a constraint that their two sub-networks share the same parameters, \ie weights and biases. As studied in our previous work~\cite{Yi-ICPR-2014}, this constraint could be removed in some conditions. Without parameters sharing, the network can deal with the view specific matching tasks more naturally. With parameters sharing, the network is more appropriate for general task, \eg cross dataset person re-identification. We call these two modes as ``General'' and ``View Specific'' SCNN. Because cross dataset problem is the main concern of this paper, we focus on General SCNN.

\subsection{Convolutional Neural Network}

\begin{figure}[!t]
\centering
\includegraphics[width=0.5\textwidth]{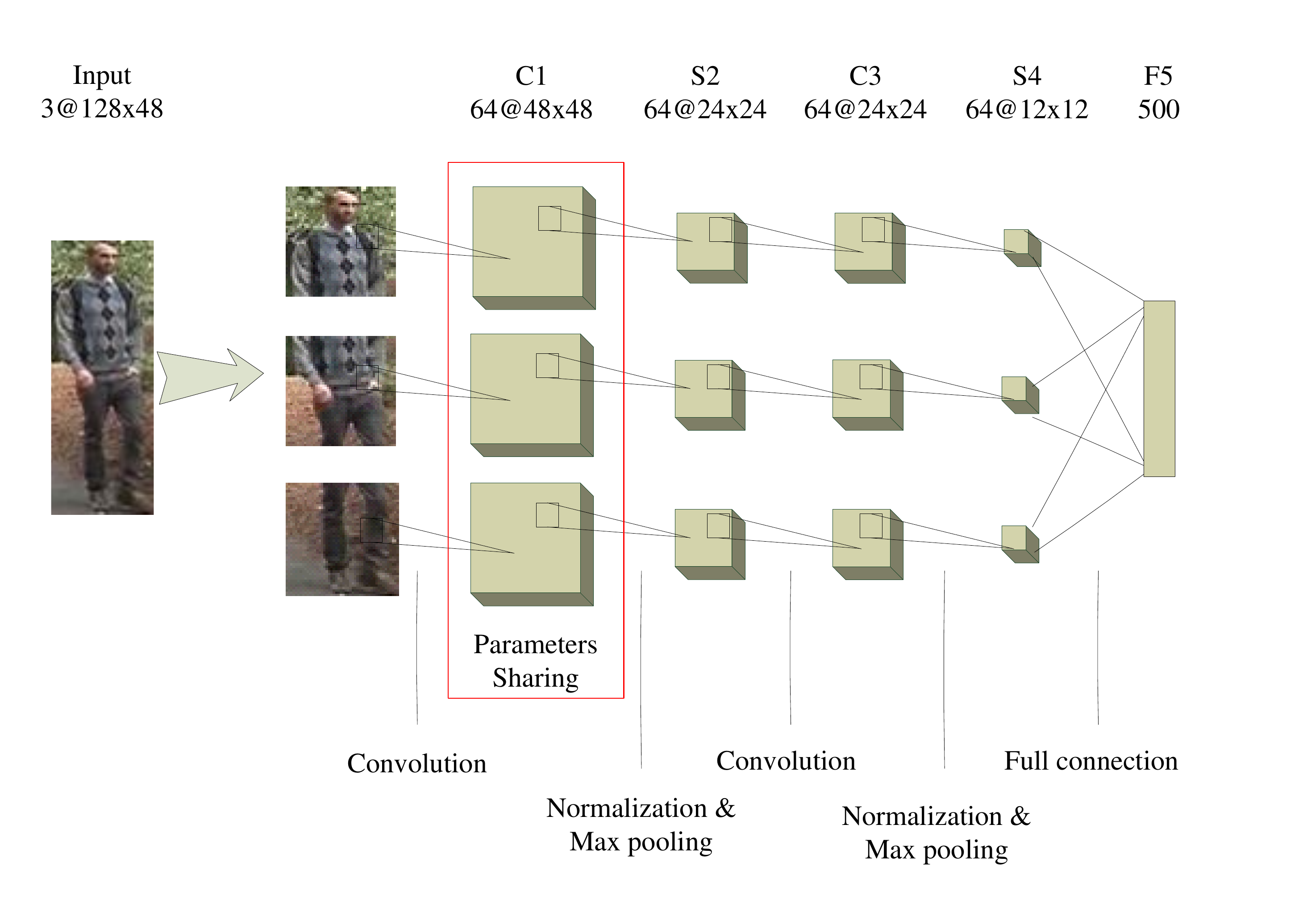}
\caption{The structure of the 5-layer CNN used in our method.}
\label{fig:cnn}
\end{figure}

The CNN in this paper (see Figure~\ref{fig:cnn}) is composed by 2 convolutional layers, 2 max pooling layers and a full connected layer. As shown in Figure~\ref{fig:cnn}, the number of channels of convolutional and pooling layers are both 64. The output of the CNN is a vector of 500 dimensions. Every pooling layer includes a cross-channel normalization unit. Before convolution the input data are padded by zero values, therefore the output has the same size with input. The filter size of C1 layer is $7\times7$ and the filter size of C2 layer is $5\times5$. ReLU neuron~\cite{Krizhevsky-NIPS-2012} is used as activation function for each layer.

To capture the different statistical properties of body parts, we train the CNN in part based way. In our previous work~\cite{Yi-ICPR-2014}, person images are cropped into three overlapped parts and three networks are trained independently. Differently, we use a faster scheme in this paper that the three parts are trained jointly. First, the three parts share C1 layer. Second, each part has its own C3 layer, which can help to learn part-specific filters. Third, the high level features of all parts are fused at F5 layer by sum rule. Then, the similarity of fused features are evaluated by the connection function. Driven by a common cost function, the three parts can contribute to the training process jointly.

Overall, there are two main differences between the proposed network in Figure~\ref{fig:cnn} and the network in \cite{Yi-ICPR-2014}: 1) C1 layer are shared by three parts or not; 2) the contribution of the three parts are fused in feature level or score level. Parameter sharing in low level can reduce the complexity of the network. Fusion in feature level make the three parts can train jointly, which will improve the performance slightly. Moreover, training and test a single network is more convenient and efficient than using three independent networks.

\subsection{Cost Function and Learning}
\label{subsec:learning}

Before learning the parameters of SCNN, we revisit its structure again. As shown in Figure~\ref{fig:arch}, the structure of SCNN can be abstracted into three basic components: two sub-networks, a connection function and a cost function. Connection function is used to evaluate the relationship between two samples and cost function is used to convert the relationship into a cost. How to choose the connection function and cost function is closely related to the performance of SCNN.

There are many distance, similarity, or other functions can be used as candidates to connect two vectors, such as: Euclidian distance, Cosine similarity, absolute difference, vector concatenate and so on. Their formulas are
\begin{eqnarray}
  S_{euc}(\bfx, \bfy) &=& -\sum_i(\bfx_i - \bfy_i)^2,  \\
  S_{cos}(\bfx, \bfy) &=& \frac{\sum_i \bfx_i\bfy_i}{\sqrt{\sum_i \bfx_i\bfx_i \sum_i \bfy_i\bfy_i}},
  \label{equ:Cosine} \\
  S_{abs}(\bfx, \bfy) &=& -\sum_i |\bfx_i - \bfy_i|, \\
  S_{con}(\bfx, \bfy) &=& \sum_i w_i [\bfx; \bfy]_i.
  \label{equ:connection}
\end{eqnarray}
In the above equations, we negate the distance functions to make them consistent to similarity. The advantage of Euclidian distance is that its derivation has simple form, but its output is unbounded which could make the training process unstable. Absolute difference is non-derivable at some points. Cosine function is bounded to [-1, 1] and invariant to the magnitude of samples. Because of the good property of Cosine function and it has been used widely in many pattern recognition problems~\cite{Bromley-NIPS-1993, Qin-IPM-2008, Nguyen-ACCV-2010}, we choose it as the connection function.

For the cost function, \cite{Yi-ICPR-2014} has given some analysis about Square loss, Exponential loss, and Binomial deviance~\cite{Friedman-Book-2009} and chose Binomial deviance as the final cost function. Here we discuss another popular cost function in pattern recognition: Fisher criterion. Given a training set $X=[\bfx_1, \bfx_2, \cdots, \bfx_n]$, its corresponding similarity matrix $S$ and mask matrix $M$, Binomial deviance and Fisher criterion are formulated as
\begin{eqnarray}
    J_{dev} &=& \sum_{i,j} W \circ \ln(e^{-\alpha (S - \beta) \circ M} + 1),
\label{equ:dev-cost} \\
    J_{fisher} &=& -\frac{(\sum_{i,j} P \circ S)^2}{\sum_{i,j} (S - \bar{S})^2},
\label{equ:fisher-cost}
\end{eqnarray}
where $\circ$ is element-wise matrix product.
\begin{eqnarray}
    S = [S_{ij}]_{n \times n}, & S_{ij}=S(\bfx_i, \bfx_j),\\
    M = [M_{ij}]_{n \times n}, & M_{ij} =\left\{\begin{array}{ll}
                                                1,  \text{positive pair} \\
                                                -1, \text{negative pair} \\
                                                0,  \text{neglected pair}
                                              \end{array}
    \right. ,\\
    W = [W_{ij}]_{n \times n}, & W_{ij} =\left\{\begin{array}{ll}
                                                \frac{1}{n_1}, \text{positive pair} \\
                                                \frac{1}{n_2}, \text{negative pair} \\
                                                0,              \text{neglected pair}
                                              \end{array}
    \right. ,\\
    P = [P_{ij}]_{n \times n}, & P_{ij} =\left\{\begin{array}{ll}
                                                \frac{1}{n_1}, \text{positive pair} \\
                                                -\frac{1}{n_2}, \text{negative pair} \\
                                                0,             \text{neglected pair}
                                              \end{array}
    \right. .
\end{eqnarray}
$S_{ij}$ denotes the similarity of sample $\bfx_i$ and $\bfx_j$. $M_{ij}$ denotes whether $\bfx_i$ and $\bfx_j$ come from the same subject or not. $n_1$ is the count of positive pairs. $n_2$ is the count of negative pairs. $\bar{S}$ is the mean of $S$. $\alpha$ and $\beta$ are hyper-parameters of Binomial deviance. The numerator of Eqn.~(\ref{equ:fisher-cost}) is the between class divergence of similarity matrix $S$ and the denominator is the total variance. By minimizing Eqn.~(\ref{equ:dev-cost}) or Eqn.~(\ref{equ:fisher-cost}), we can learn a network separating the positive and negative pairs as far as possible.

By comparing Eqn.~(\ref{equ:dev-cost}) and Eqn.~(\ref{equ:fisher-cost}), we can see that Binomial deviance cost more focus on the false classified samples (or the samples near the boundary), but Fisher criterion focus on every elements of the similarity matrix $S$ equally. In our intuition, Binomial deviance cost can make the network be trained mainly on the hard samples and more likely to get a good model, which will be verified in the experiments. In the following sections, we fix the connection function as Cosine and the cost function as Binomial deviance.

After the connection and cost functions are determined, Back-Propagation (BP)~\cite{LeCun-IEEE-1998} is used to learn the parameters of SCNN. By plugging Eqn.~(\ref{equ:Cosine}) into Eqn.~(\ref{equ:dev-cost}), we can get the forward propagation function to calculate the cost of training set.
\begin{equation}
\label{equ:fp}
    J_{dev}(X) = \sum_{i,j} W \circ \ln(e^{-\alpha (S - \beta) \circ M} + 1),
\end{equation}
where $X$ is the output of CNN. $S$ is the output of connection function.
\begin{equation}
S = [S_{ij}]_{n \times n}, S_{ij}=\frac{\bfx_i^T \bfx_j}{\sqrt{\bfx_i^T \bfx_i \bfx_j^T \bfx_j}}.
\end{equation}
Differentiating the cost function with respect to $X$, we can get
\begin{equation}
\label{equ:bp-x}
  \frac{\partial J_{dev}}{\partial X} = X (AB + (AB)^T) - X \circ (\widetilde{AC} + \widehat{AD}),
\end{equation}
where
\begin{eqnarray}
A = -\alpha W \circ M \circ \frac{e^{-\alpha (S - \beta) \circ M}}{e^{-\alpha (S - \beta) \circ M} + 1}, \\
\label{equ:B}
B = [B_{ij}]_{n \times n}, B_{ij}=\frac{1}{\sqrt{\bfx_i^T \bfx_i \bfx_j^T \bfx_j}},\\
\label{equ:C}
C = [C_{ij}]_{n \times n}, C_{ij}=B_{ij} \frac{\bfx_i^T \bfx_j}{\bfx_i^T \bfx_i}, \\
\label{equ:D}
D = [D_{ij}]_{n \times n}, D_{ij}=B_{ij} \frac{\bfx_i^T \bfx_j}{\bfx_j^T \bfx_j}, \\
\label{equ:AC}
\widetilde{AC} = repmat(\left(\sum_j A_{ij} C_{ij} \right)^T, d, 1), \\
\label{equ:AD}
\widehat{AD} = repmat(\sum_i A_{ij} D_{ij}, d, 1).
\end{eqnarray}
where ``repmat'' is a function to create a matrix by tiling vector many times, as same as the function in Matlab.
One can refer Appendix~\ref{appdx:general} for the detail derivation of Eqn.~(\ref{equ:bp-x}).

While \cite{Yi-ICPR-2014} trains the network in a pairwise way, this paper formulates the cost and gradient in totally matrix form. When the network is trained by Stochastic Gradient Descent (SGD) with mini-batch, the new formulation can process more sample pairs in one batch. When the size of batch is set to 128, in \cite{Yi-ICPR-2014} a batch includes 64 positive and 64 negative sample pairs, but in this paper a batch can generate $128 \times (128 - 1) / 2 = 8128$ pairs, which makes the network scans the training data faster and saves the training time.

Based on Eqn.~(\ref{equ:fp}) and Eqn.~(\ref{equ:bp-x}), we can learn the parameters of SCNN by SGD algorithm. For general neural network, the error is backward propagated from top to down through a single path. On the contrary, the error of specific SCNN is backward propagated through two branches by Eqn.~(\ref{equ:bp-x1}) and Eqn.~(\ref{equ:bp-y1}) respectively as described in \cite{Yi-ICPR-2014}. In practice, we also can assign asymmetry cost on the label $M_{ij}$ to positive and negative pairs to tune the performance of network, \eg $1$ for positive pairs and $-2$ for negative pairs. The effect of asymmetry cost on the performance will be discussed in the experiments.

\section{Experiments}


There are many popular datasets were built for person re-identification, such as VIPeR~\cite{Gray-PETS-2007}, PRID 2011~\cite{Hirzer-IA-2011}, i-LIDS~\cite{Zheng-CVPR-2011}, CUHK Campus~\cite{Li-CVPR-2013} and so on. Among these datasets, the evaluation protocols of VIPeR and PRID (single-shot version) are the clearest two, therefore we compare our method with other methods on them. The experiments are done in two settings:
\begin{enumerate}
\item Intra dataset experiments
    \begin{itemize}
    \item training and test both on VIPeR;
    \item training and test both on PRID.
    \end{itemize}
\item Cross dataset experiments
    \begin{itemize}
    \item training on i-LIDS and test on VIPeR;
    \item training on i-LIDS and test on PRID;
    \item training on CUHK Campus and test on VIPeR;
    \item training on CUHK Campus and test on PRID.
    \end{itemize}
\end{enumerate}

The intra dataset experiments are conducted to illustrate the basic performance of the proposed method. The cross dataset experiments are to illustrate the generalization ability.

\subsection{Intra Dataset}

Except for \cite{Ma-ICCV-2013} and \cite{Yi-ICPR-2014}, other papers all conduct experiments in this setting that is training and test on the same dataset. VIPeR includes 632 subjects, 2 images per subject coming from 2 different camera views (camera A and camera B). We split VIPeR into disjoint training (316 subjects) and testing set (316 subjects) randomly, and repeat the process 11 times. The first split (Dev. view) is used for parameter tuning, such as the number of training epoch, learning rate, weight decay and so on. The other 10 splits (Test view) are used for reporting the results.

As similar to VIPeR, PRID is captured by 2 cameras too (camera A and camera B). Camera A shows 385 subjects and camera B shows 749 subjects. The first 200 subjects appear in both cameras. Follow the testing protocols in \cite{Hirzer-IA-2011, Hirzer-ECCV-2012}, we randomly select 100 subjects from the first 200 subjects for training and the remain subjects for testing. The test set is composed by probe and gallery, and their information are as follows.
\begin{itemize}
\item Probe set: the remain 100 subjects of the first 200 subjects in camera A except for the 100 training subjects;
\item Gallery set: the remain 649 subjects of camera B except for the 100 training subjects.
\end{itemize}
Again, the whole process is repeated 11 times. The first split is used for parameter tuning and the other 10 splits are used for reporting the results.

In the training stage, all training images from camera A and camera B are merged, randomly shuffled and sent to a generic SCNN. The corresponding mask matrix is generated according to the label of samples. Those pairs from the same subjects are assigned 1, and those from the different subjects are assigned -1. Because the mask matrix is symmetric, we set the elements in the lower triangular part to 0 to avoid redundant computation. In the testing stage, one image of each subject is used as gallery and the other one is used as probe.

Before evaluate on Test view, we use Dev. view to investigate the most three important factors which affect the performance of network: 1) data augmentation, 2) asymmetric cost for positive and negative sample pairs, and 3) cost function. Besides, we set the parameters of the cost function Eqn.~(\ref{equ:dev-cost}) as $\alpha=2$ and $\beta=0.5$.

\subsubsection{Data Augmentation}

Data augmentation is a widely used trick~\cite{Cirecsan-AAAI-2011} for training neural network. In the experiments, we will mirror all person images to double the training and test sets. Although this trick has been used in \cite{Yi-ICPR-2014}, but its effect on the performance was not analyzed. Here, we compare the performance on Dev. view of VIPeR with and without data augmentation. For the training set, the number of images of each subject is increased from 2 to 4. In the testing stage, 2 original images and their mirrored version generate 4 similarity scores and the final score are fused by sum rule.

\begin{table*}[!t]
\caption{The rank-1 to rank-50 recognition rates on Dev. view of VIPeR with or without data augmentation.}
\label{tbl:data-aug}
\centering
\begin{tabular}{|c||c|c|c|c|c|c|c|c|}
\hline
Rank & 1 & 5 & 10 & 15 & 20 & 25 & 30 &50\\
\hline
With Aug. & 34.49\% & 60.13\% & 74.37\% & 80.7\% & 84.18\% & 88.61\% & 91.14\% & 96.84\% \\
\hline
Without Aug. & 18.4\% & 49.37\% & 67.09\% & 74.05\% & 80.06\% & 84.18\% & 87.04\% & 96.2\%\\
\hline
\end{tabular}
\end{table*}

From the results in Table~\ref{tbl:data-aug}, we can see significant improvements brought by data augmentation, especially in the top ranks. This indicates that the scale of dataset is crucial to train good networks. If know the geometry information of the person in the image, \ie the 3D pose of the person, we can augment the datasets guided by the 3D pose and generate more virtual images to improve the performance. In this sense, human 3D pose estimation will be an important direction for person re-identification in the following years.

\subsubsection{Asymmetric Cost}

As described in Section~\ref{subsec:learning}, while generating mask matrix from the labels of training samples, the number of negative sample pairs is far more than positive pairs. However, in practice we usually split the training samples into many batches first, and then the positive and negative sample pairs are just generated within each batch. Therefore, the negative sample pairs between batches can not be covered by the training process. This may cause the negative pairs prone to under-fitting.

To balance the weight of positive and negative sample pairs, we can assign asymmetric costs to them. While fixing the cost of positive pair to $1$, we tune the cost of negative pair $c$ from $1$ to $3$. The asymmetric cost can apply easily on Eqn.~(\ref{equ:fp}) and Eqn.~(\ref{equ:bp-x}) by setting
\begin{equation}
\label{equ:asym-cost}
    M_{ij} =\left\{\begin{array}{ll}
                                                1,  \text{positive pair} \\
                                                -c, \text{negative pair} \\
                                                0,  \text{neglected pair}
                                              \end{array}
    \right.
\end{equation}

Table~\ref{tbl:asym-cost} shows the relationship between the negative cost $c$ and recognition rate. On Dev. view of VIPeR, the highest overall performance is achieved at $c=2$. This illustrates that the negative pairs should be paid more attention in each training batch.

\begin{table*}[!t]
\caption{The recognition rate on Dev. view of VIPeR at different negative costs.}
\label{tbl:asym-cost}
\centering
\begin{tabular}{|c||c|c|c|c|c|c|c|c|}
\hline
Rank & 1 & 5 & 10 & 15 & 20 & 25 & 30 & 50\\
\hline
c = 1 & 26.9\% & 53.48\% & 65.82\% & 74.05\% & 78.48\% & 85.76\% & 87.34\% & 94.94\% \\
\hline
c = 1.5 & 31.01\% & 58.54\% & 70.89\% & 78.8\% & 84.49\% & 87.66\% & 90.82\% & 97.78\% \\
\hline
c = 2 & 34.49\% & 60.13\% & 74.37\% & 80.7\% & 84.18\% & 88.61\% & 91.14\% & 96.84\% \\
\hline
c = 2.5 & 29.75\% & 59.18\% & 71.84\% & 80.06\% & 85.13\% & 89.24\% & 91.77\% & 96.2\% \\
\hline
c = 3 & 27.85\% & 56.65\% & 70.57\% & 78.8\% & 85.44\% & 87.97\% & 91.46\% & 94.94\% \\
\hline
\end{tabular}
\end{table*}

\subsubsection{Binomial Deviance vs. Fisher Criterion}

Next, we compare two cost functions discussed in Section~\ref{subsec:learning}: Binomial deviance (Eqn.~(\ref{equ:dev-cost})) and Fisher criterion (Eqn.~(\ref{equ:fisher-cost})). Like the two prior two experiments, the comparison is conducted on Dev. view of VIPeR too. In the comparison experiments, we mirror the images to double the dataset and set the cost for negative pairs to $c=2$. The connection function is fixed to Cosine. The differences between Binomial deviance and Fisher criterion are evaluated in three aspects: 1) epoch-cost curve, 2) similarity score distribution, and 3) recognition rate.

\begin{figure}
  \centering
  \includegraphics[width=0.24\textwidth]{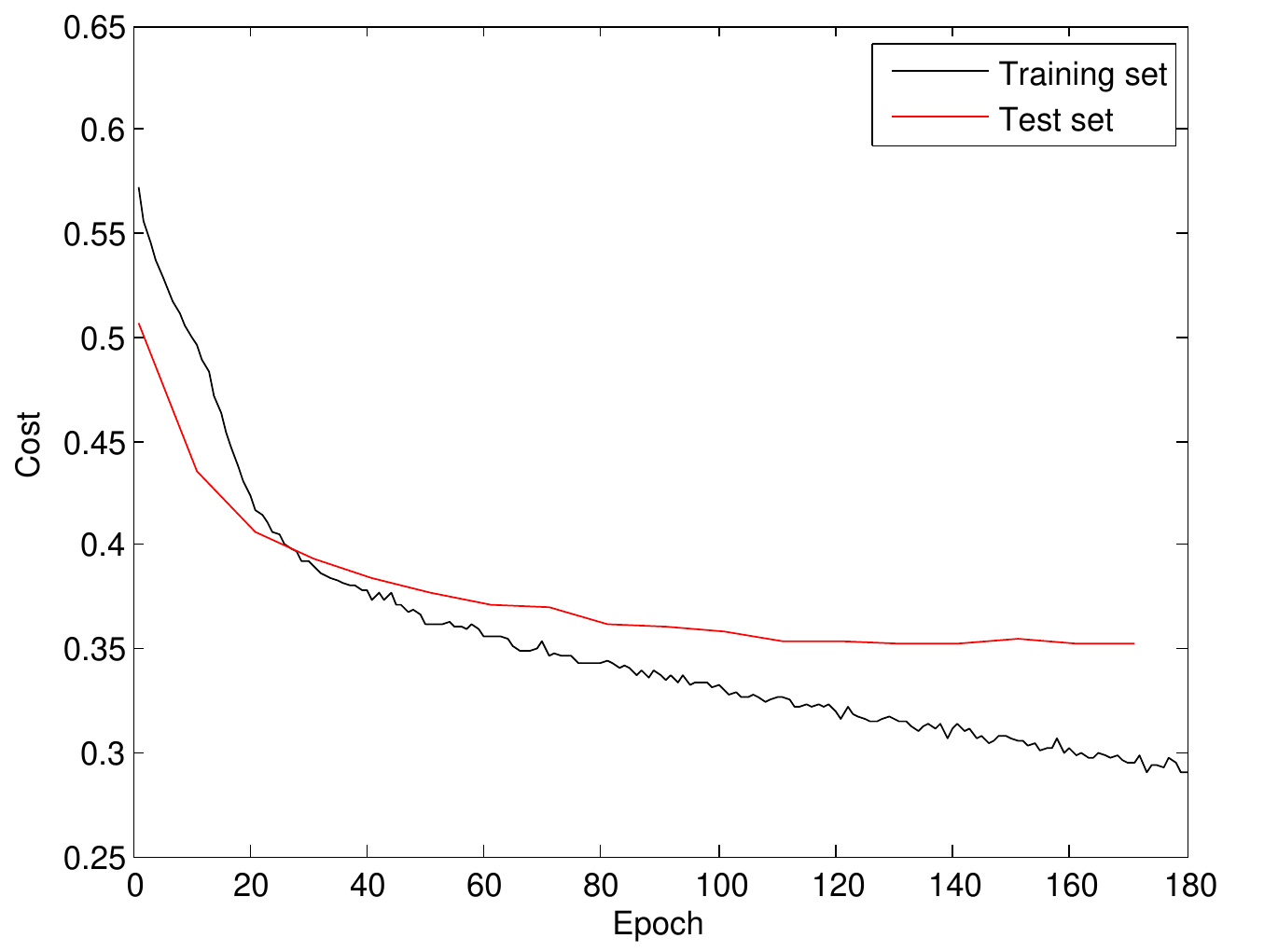}
  \includegraphics[width=0.24\textwidth]{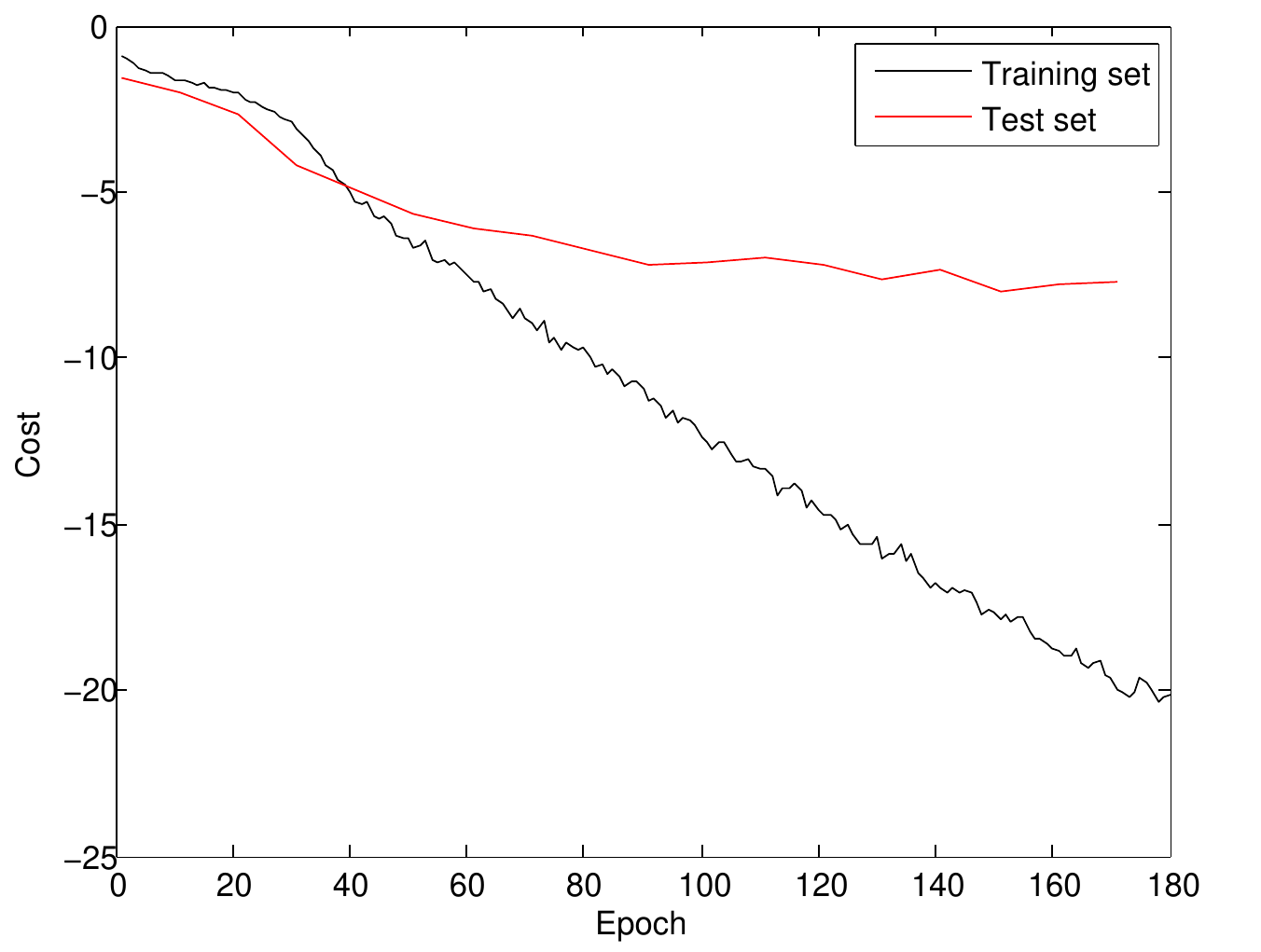}\\
  \caption{The epoch-cost curves of the training and test set (Dev. view of VIPeR) when using Binomial deviance (left) or Fisher criterion (right) as cost function.}
  \label{fig:epoch-cost}
\end{figure}

\begin{figure}
  \centering
  \includegraphics[width=0.5\textwidth]{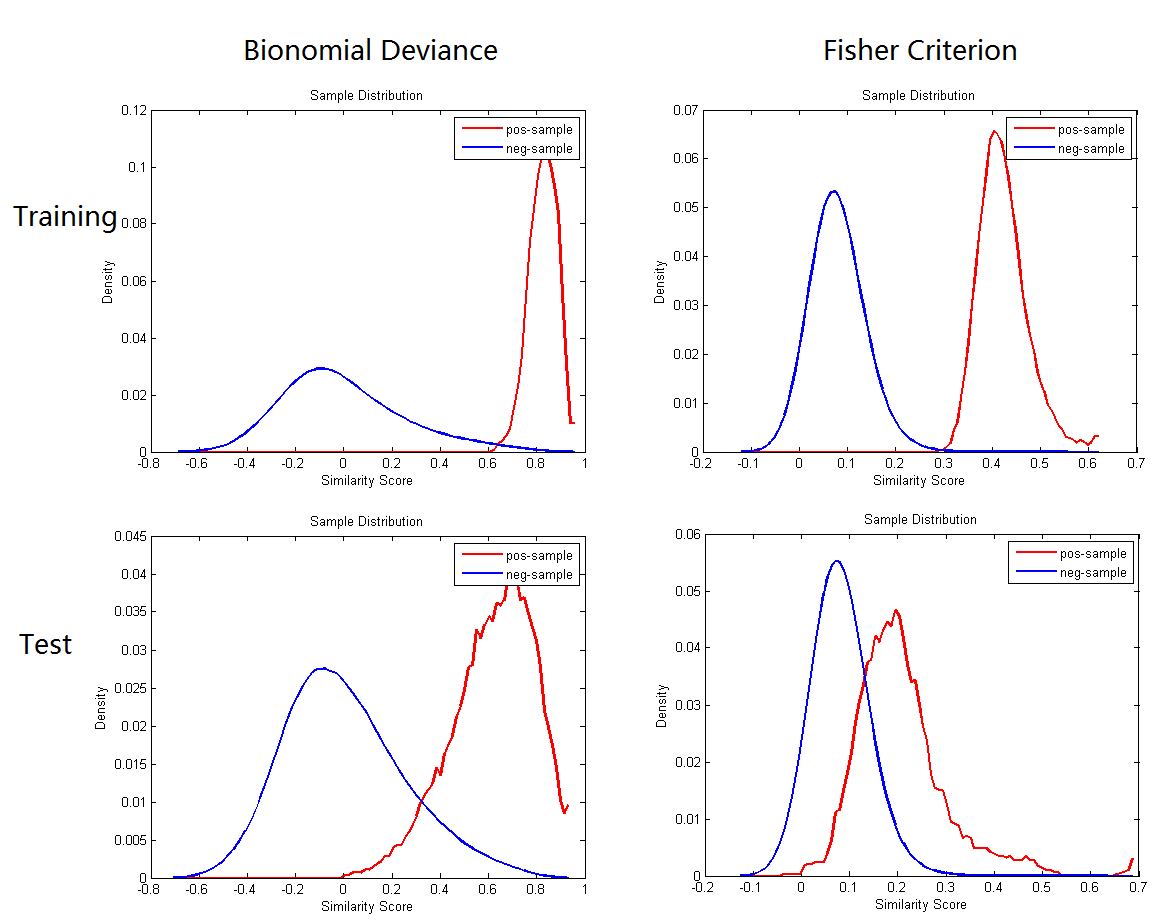}
  \caption{The similarity score distributions of the training and test set (Dev. view of VIPeR) when using Binomial deviance (left) or Fisher criterion (right) as cost function.}
  \label{fig:distribution}
\end{figure}

\begin{figure}
  \centering
  \includegraphics[width=0.25\textwidth]{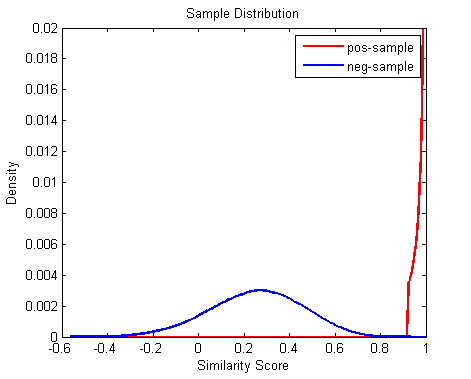}
  \caption{The similarity score distribution of VIPeR computed based on the 21 human labelled attributes~\cite{Layne-PRI-2014}. The 21 attributes are binary and the similarity scores are evaluated by Hamming distance.}
  \label{fig:attr}
\end{figure}

\begin{table*}[!t]
\caption{The rank-1 to rank-50 recognition rates on Dev. view of VIPeR using different cost functions.}
\label{tbl:cost-function}
\centering
\begin{tabular}{|c||c|c|c|c|c|c|c|c|}
\hline
Rank & 1 & 5 & 10 & 15 & 20 & 25 & 30 &50\\
\hline
Binomial Deviance & 34.49\% & 60.13\% & 74.37\% & 80.7\% & 84.18\% & 88.61\% & 91.14\% & 96.84\% \\
\hline
Fisher Criterion & 14.24\% & 35.13\% & 47.15\% & 56.96\% & 62.66\% & 67.41\% & 71.84\% & 80.06\%\\
\hline
\end{tabular}
\end{table*}

Figure~\ref{fig:epoch-cost} shows the epoch-cost curves on Dev. view of VIPeR. Low cost reflects high performance approximately, although there is no explicit relationship between the cost and the recognition rate. No matter using which cost function, the cost of the training set always drop continually. On the contrary, the cost of the test set drops significantly just at the beginning of training process, and then it gradually becomes converged after tens of epochs. From the figure we can see that the cost gap between the training and test set is small when using Binomial deviance. But for Fisher criterion, the gap is obviously bigger than Binomial deviance, which reflects that Fisher criterion is easily overfiting to the training set. By inspecting the epoch-cost curves, we set epoch$=180$ based on our experience in the following experiments.

As shown in Figure~\ref{fig:distribution}, the two-class similarity distributions of the two cost functions are very different. And a distribution generated from the attributes~\cite{Layne-PRI-2014} of VIPeR is also given in Figure~\ref{fig:attr} for reference. The experiment in \cite{Layne-PRI-2014} has shown that only using the 21 attributes and Hamming distance can achieve very high recognition rate, \ie rank1$\approx76\%$. Because the attributes are labelled by human, they can be seen as a baseline of human performance, and the distribution in Figure~\ref{fig:attr} can be seen as the ideal two-class distribution of VIPeR. For Binomial deviance, the distribution of negative similarity scores are significantly wider than that of positive scores, which is very coincide with the ideal distribution generated from the 21 attributes. For Fisher criterion, the distributions of positive and negative scores are standard Gaussian with the same variance.

Although nearly perfect results can be obtained on the training set using any cost function, their performance on the test set are different (see Table~\ref{tbl:cost-function}). From the performance, we can see that Binomial deviance is more suitable for person re-identification problem because it mainly focus on the samples near the boundary and less affected by the distributions of positive and negative samples. For Fisher criterion, the ideal distributions are highly heteroscedastic, which is conflicted with the assumption of Fisher. Maybe some modifications can be made on Fisher criterion to solve this problem, but we leave the work to future.

\subsubsection{Results}

After the above analysis and parameter tuning, we test the performance of our network on Test View of VIPeR and PRID with the following configuration:
\begin{itemize}
\item Augment the training and set set by mirror the samples and fused the score by sum;
\item Set the negative cost $c=2$; set the number of epoch$=180$;
\item Use Cosine as connection function and Binomial Deviance as cost function.
\end{itemize}
The experiments are repeated 10 times and the mean recognition rates are list in Table~\ref{tbl:VIPeR-results} and Table~\ref{tbl:PRID-results}.

\begin{table*}[!t]
\caption{Intra dataset experiment: comparison of the proposed method and other state-of-the-art methods on VIPeR.}
\label{tbl:VIPeR-results}
\centering
\begin{tabular}{|c||c|c|c|c|c|c|c|c|}
\hline
\backslashbox{Method}{Rank} & 1 & 5 & 10 & 15 & 20 & 25 & 30 & 50\\
\hline
ELF~\cite{Gray-ECCV-2008} & 12\% & 31\% & 41\% & - & 58\% & - & - & - \\
\hline
RDC~\cite{Zheng-TPAMI-2013} & 15.66\% & 38.42\% & 53.86\% & - & 70.09\% & - & - & - \\
\hline
PPCA~\cite{Mignon-CVPR-2012} & 19.27\% & 48.89\% & 64.91\% & - & 80.28\% & - & - & - \\
\hline
Salience~\cite{Zhao-CVPR-2013} & 26.74\% & 50.7\% & 62.37\% & - & 76.36\% & - & - & - \\
\hline
RPML~\cite{Hirzer-ECCV-2012} & 27\% & - & 69\% & - & 83\% & - & - & 95\% \\
\hline
LAFT~\cite{Li-CVPR-2013} & 29.6\% & - & 69.31\% & - & - & 88.7\% & - & {\bf 96.8}\% \\
\hline
Salience1~\cite{Zhao-ICCV-2013} & 30.16\% & 52.3\% & - & - & - & - & - & - \\
\hline
DML~\cite{Yi-ICPR-2014} & 28.23\% & 59.27\% & 73.45\% & 81.2\% & 86.39\% & 89.53\% & 92.28\% & 96.68\% \\
\hline
Improved DML & {\bf 34.4}\% & {\bf 62.15}\% & {\bf 75.89}\% & {\bf 82.56}\% & {\bf 87.22}\% & {\bf 89.65}\% & {\bf 92.28}\% & 96.52\% \\
\hline
\end{tabular}
\end{table*}

To distinguish the modified DML from the original one in \cite{Yi-ICPR-2014}, we rename the proposed method in this paper as ``Improved DML''. Compared to the pairwise version in \cite{Yi-ICPR-2014}, the improved DML has 5$\times$ training speed and significant performance improvement. Notably, the rank-1 recognition rate increases by 6\%, from 28.23\% to 34.4\%. The results of other compared methods are copied from the original papers. If the results are unavailable, they are leaved as ``-''. From the table we can see that the proposed method outperforms most of compared methods on VIPeR including the current state-of-the-art methods~\cite{Li-CVPR-2013, Zhao-ICCV-2013}. From rank-1 to rank-50, our method outperforms \cite{Li-CVPR-2013, Zhao-ICCV-2013} remarkably, except that the rank-50 recognition rate of \cite{Li-CVPR-2013} is higher than ours slightly. Among all methods, our method is nearly the most simple and elegant one, which doesn't need any pre-processing, pose information or segmentation. From the bottom to top layers in the network, every building blocks contribute to a common objective function and are optimized by BP algorithm simultaneously.

\begin{table*}[!t]
\caption{Intra dataset experiment: comparison of the proposed method and other state-of-the-art methods on PRID 2011.}
\label{tbl:PRID-results}
\centering
\begin{tabular}{|c||c|c|c|c|c|c|c|c|}
\hline
\backslashbox{Method}{Rank} & 1 & 5 & 10 & 15 & 20 & 25 & 30 & 50\\
\hline
Descr. Model~\cite{Hirzer-IA-2011} & 4\% & - & 24\% & - & 37\% & - & - & 56\% \\
\hline
RPML~\cite{Hirzer-ECCV-2012} & 15\% & - & 42\% & - & 54\% & - & - & 70\% \\
\hline
Improved DML & {\bf 17.9\%} & {\bf 37.5}\% & {\bf 45.9}\% & {\bf 50.7}\% & {\bf 55.4}\% & {\bf 59.3}\% & {\bf 63.1}\% & {\bf 71.4}\% \\
\hline
\end{tabular}
\end{table*}

On PRID, we get similar results to VIPeR that the proposed method outperforms the state-of-the art method, RPML~\cite{Hirzer-ECCV-2012}. However, the superiority of our method on PRID (1\% to 3\%) is smaller than that on VIPeR (4\% to 10\%), the reason is that the scale of PRID dataset is too small to train a good network. Just using 100 sample pairs, our network outperforms the compared methods, which can illustrate the power of deep metric learning. Compared to VIPeR, the quality of PRID is poorer and the size of its gallery is bigger, therefore the recognition rates on PRID are overall lower than VIPeR.

\subsection{Cross Dataset}

In this section, we conduct many experiments in the cross dataset setting which is more coincide with practical applications. In the following, i-LIDS, CUHK Campus are used as training set, and VIPeR, PRID are used as test set. i-LIDS is captured at a airport in indoor environment; CUHK Campus is captured in a campus; VIPeR and PRID are captured on the street. Due to totally different capture environments and devices, cross dataset experiments are more challenging than the previous experiments. Different from \cite{Ma-ICCV-2013}, we don't use any samples in VIPeR and PRID to adapt classifiers to the target domains.

In i-LIDS dataset, there are 119 people with total 476 images captured by multiple non-overlapping cameras with an average of 4 images for each person. Many of these images have large illumination changes and are subject to occlusions. The scale of i-LIDS is smaller than VIPeR and PRID and the resolution of images in i-LIDS are varied. CUHK Campus is a large scale dataset captured by two cameras in 5 sessions, includes 1816 subjects and 7264 images. Each subject has 4 images from 2 camera views. The resolution of CUHK Campus is $60 \times 160$. The cross dataset networks are trained with the same parameters as the previous experiments.

Before training, we resize all images in i-LIDS and CUHK Campus to $40 \times 128$ for convenience. For the test set, we use the same setting with the intra dataset experiments. For VIPeR, A half of subjects and images are randomly selected to construct the test set, which includes 316 subjects and 632 images. For PRID, the test set includes 649 subjects and 749 images. The test process is repeated 10 times too, and the average recognition rate is reported.

\subsubsection{Results}

\begin{table}[!t]
\caption{Cross dataset experiment: comparison of the proposed method and DTRSVM~\cite{Ma-ICCV-2013} on VIPeR.}
\label{tbl:VIPeR-cross-results}
\centering
\begin{tabular}{|c|c|c||c|c|c|c|}
\hline
Methods & Tr. Set & 1 & 10 & 20 & 30\\
\hline
DML~\cite{Yi-ICPR-2014} & CUHK & 16.17\% & 45.82\% & 57.56\% & 64.24\% \\
\hline
DTRSVM~\cite{Ma-ICCV-2013} & PRID & 10.9\% & 28.2\% & 37.69\% & 44.87\% \\
\hline
DTRSVM~\cite{Ma-ICCV-2013} & i-LIDS & 8.26\% & 31.39\% & {\bf 44.83}\% & {\bf 53.88}\% \\
\hline
Improved DML & i-LIDS & {\bf 11.61}\% & {\bf 34.43}\% & 44.08\% & 52.69\% \\
\hline
Improved DML & CUHK & {\bf 16.27}\% & {\bf 46.27}\% & {\bf 59.94}\% & {\bf 70.13}\% \\
\hline
Improved DML & \tabincell{c}{i-LIDS\\+ CUHK} & {\bf 17.72}\% & {\bf 48.8}\% & {\bf 63.35}\% & {\bf 72.85}\% \\
\hline
\end{tabular}
\end{table}

\begin{table}[!t]
\caption{Cross dataset experiment: comparison of the proposed method and DTRSVM~\cite{Ma-ICCV-2013} on PRID 2011.}
\label{tbl:PRID-cross-results}
\centering
\begin{tabular}{|c|c|c||c|c|c|}
\hline
Methods & Tr. Set & 1 & 10 & 20 & 30\\
\hline
DTRSVM~\cite{Ma-ICCV-2013} & VIPeR & 4.6\% & 17.25\% & 22.9\% & 28.1\% \\
\hline
DTRSVM~\cite{Ma-ICCV-2013} & i-LIDS & 3.95\% & 18.85\% & 26.6\% & 33.2\% \\
\hline
Improved DML & i-LIDS & {\bf 8}\% & {\bf 25.5}\% & {\bf 38.9}\% & {\bf 45.6}\% \\
\hline
Improved DML & CUHK & {\bf 7.6}\% & {\bf 23.4}\% & {\bf 30.9}\% & {\bf 36.1}\% \\
\hline
Improved DML & \tabincell{c}{i-LIDS\\+ CUHK} & {\bf 13.8}\% & {\bf 35.4}\% & {\bf 45}\% & {\bf 51.3}\% \\
\hline
\end{tabular}
\end{table}

First, we use i-LIDS and CUHK Campus as training set respectively, and test the performance of the trained networks on VIPeR. The recognition rates on VIPeR are shown in Table~\ref{tbl:VIPeR-cross-results} and the results of DTRSVM~\cite{Ma-ICCV-2013} are listed for comparison. From the results we can see that our method outperforms the original DML in \cite{Yi-ICPR-2014} slightly. When training on i-LIDS, our method is on a par with DTRSVM, and the rank-1 and rank-10 recognition rates are better than DTRSVM slightly. When training on CUHK Campus, our performance improved significantly. A possible reason is that compared to i-LIDS the quality and aspect ratio of images in CUHK Campus is more similar to VIPeR. By combining the similarity scores of the two networks by sum, the performance is improved further and even approaches the performance of some methods in intra dataset setting, such as ELF~\cite{Gray-ECCV-2008} and RDC~\cite{Zheng-TPAMI-2013}.

Then we test the two trained networks on PRID, and give the results in Table~\ref{tbl:PRID-cross-results}. When using i-LIDS as training set, our method is better than DTRSVM significantly. But different from the results on VIPeR, training on CUHK Campus decreases the recognition rates remarkably. The performance decline is caused by the big difference between CUHK Campus and PRID datasets. Fusing the similarity scores of i-LIDS and CHUK Campus increases the performance too, especially the rank-1 recognition raises from 8\% to 13.8\%.

\begin{figure}
    \centering
\subfloat[VIPeR]{\includegraphics[width=0.3\textwidth]{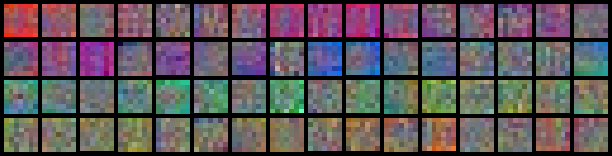}}\\
\subfloat[PRID]{\includegraphics[width=0.3\textwidth]{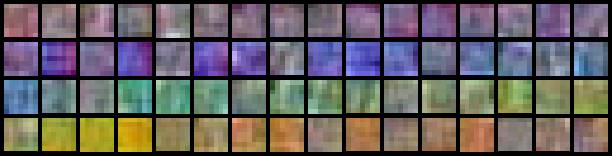}}\\
\subfloat[i-LIDS]{\includegraphics[width=0.3\textwidth]{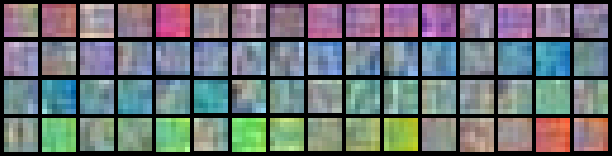}}\\
\subfloat[CUHK]{\includegraphics[width=0.3\textwidth]{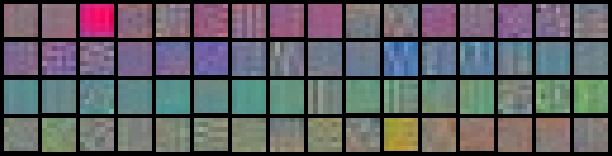}}
    \caption{The filters in the first layer of CNN, from top to down, which are learned on VIPeR, PRID, i-LIDS, CUHK Campus respectively. The size of filters are $7 \times 7$, and their order are sorted by Hue component for best view.}
    \label{fig:filters}
\end{figure}

Compared to the intra dataset experiments, the performance of cross dataset experiments decline sharply. For rank-30 recognition rate, the number of VIPeR and PRID both drop more than 20\%, which indicates the trained models are hard to generalize across datasets due to the distinctive properties of each dataset. Figure~\ref{fig:filters} shows some filters learned from different datasets, from which we can see the distributions of color and texture on these datasets are very diverse. Besides of the experimental results, the diverse filters also can prove that the model learned on a dataset can hardly adapt to another one. How to transfer a model to target domain or how to fully use the multiple heterogeneous datasets at hand to improve the performance in target domain are important research topics in the future.

\section{Conclusions}

This paper proposed a deep metric learning method by using siamese convolutional neural network. The structure of the network and the training process were described in detail. Extensive intra dataset and cross dataset person re-identification experiments were conducted to illustrate the superiorities of the proposed method. This is the first work to apply deep learning in the person re-identification problem and is also the first work to study the person re-identification problem in fully cross dataset setting. The experimental results illustrated that the network can deal with the cross view and cross dataset person re-identification problems efficiently and outperformed the state-of-the-art methods significantly. In the future, we will apply DML to other applications; explore the way to pre-train the network; and investigate how to embed geometry information into the network to improve the robustness to pose variations. Moreover, we will continue to research how to train a general person matching engine with good generalization across view and dataset.

\appendices
\section{Gradients of General SCNN}
\label{appdx:general}

For general SCNN, the two sub-networks share their parameters, therefore they have the same output for a sample. Input a training set, we denote the output of CNN by $X=[\bfx_1, \bfx_2, \cdots, \bfx_n]$, where $n$ is the number of samples. The cost produced by the training set $X$ can be calculated by Eqn.~(\ref{equ:fp}). The gradient of Eqn.~(\ref{equ:fp}) with respect to $X$ can be derived as follows.
\begin{equation}
\begin{aligned}
\label{equ:jd}
\frac{\partial J_{dev}}{\partial X} = & \sum_{ij} \frac{\partial J_{ij}}{\partial X} \\
                                      = & \sum_{ij} -\alpha W \circ M \circ \frac{e^{-\alpha (S - \beta) \circ M}}{e^{-\alpha (S - \beta) \circ M} + 1} \frac{\partial S_{ij}}{\partial X} \\
                                      = & \sum_{ij} A_{ij} \frac{\partial S_{ij}}{\partial X},
\end{aligned}
\end{equation}
where $A_ij$ can be seen as a weight for each sample pair. And $\frac{\partial S_{ij}}{\partial X}$ is the derivative of the similarity $S_{ij}$ with respect to $X$.

Expand the Cosine similarity $S_{ij}$ by Eqn.~(\ref{equ:Cosine}), we can get
\begin{equation}
\begin{aligned}
\label{equ:sd}
\frac{\partial S_{ij}}{\partial X} = & \frac{\sqrt{\bfx_i^T \bfx_i \bfx_j^T \bfx_j} \frac{\partial \bfx_i^T \bfx_j}{\partial X} - \bfx_i^T \bfx_j \frac{\partial \sqrt{\bfx_i^T \bfx_i \bfx_j^T \bfx_j}}{\partial X}}{\bfx_i^T \bfx_i \bfx_j^T \bfx_j} \\
                                   = & \frac{1}{\sqrt{\bfx_i^T \bfx_i \bfx_j^T \bfx_j}} \\
& \left(
\frac{\partial \bfx_i^T \bfx_j}{\partial X} -
\frac{\bfx_i^T \bfx_j}{2\bfx_i^T \bfx_i} \frac{\partial \bfx_i^T \bfx_i}{\partial X} -
\frac{\bfx_i^T \bfx_j}{2\bfx_j^T \bfx_j} \frac{\partial \bfx_j^T \bfx_j}{\partial X} \right) \\
                                   = & B_{ij} \frac{\partial \bfx_i^T \bfx_j}{\partial X}
                                   - \frac{C_{ij}}{2} \frac{\partial \bfx_i^T \bfx_i}{\partial X}
                                   - \frac{D_{ij}}{2} \frac{\partial \bfx_j^T \bfx_j}{\partial X},
\end{aligned}
\end{equation}
where $B_{ij}$, $C_{ij}$ and $D_{ij}$ are defined in Eqn.~(\ref{equ:B}), Eqn.~(\ref{equ:C}) and Eqn.~(\ref{equ:D}) respectively. The derivatives of $\bfx_i^T \bfx_j$, $\bfx_i^T \bfx_i$ and $\bfx_j^T \bfx_j$ with respect to $X$ are as follows.
\begin{equation}
\label{equ:xx1}
\frac{\partial \bfx_i^T \bfx_j}{\partial X} = \bordermatrix{
                                                    ~ & & i & & j & \cr
                                                      &0&x_j&0&x_i&0\cr}
\end{equation}
\begin{equation}
\label{equ:xx2}
\frac{\partial \bfx_i^T \bfx_i}{\partial X} = \bordermatrix{
                                                    ~ & & i & \cr
                                                    2 &0&x_i&0\cr}
\end{equation}
\begin{equation}
\label{equ:xx3}
\frac{\partial \bfx_j^T \bfx_j}{\partial X} = \bordermatrix{
                                                    ~ & & j & \cr
                                                    2 &0&x_j&0\cr}
\end{equation}

By substituting Eqn.~(\ref{equ:xx1}), Eqn.~(\ref{equ:xx2}) and Eqn.~(\ref{equ:xx3}) into Eqn.~(\ref{equ:sd}) and substituting Eqn.~(\ref{equ:sd}) into Eqn.~(\ref{equ:jd}), we can get the final formulation of $\frac{\partial J_{dev}}{\partial X}$.
\begin{equation}
\begin{aligned}
  \frac{\partial J_{dev}}{\partial X} = & \sum_{ij} A_{ij} B_{ij} \bordermatrix{
                                                    ~ & & i & & j & \cr
                                                      &0&x_j&0&x_i&0\cr} \\
                                        - & \sum_{ij} A_{ij} C_{ij} \bordermatrix{
                                                    ~ & & i & \cr
                                                      &0&x_i&0\cr} \\
                                        - & \sum_{ij} A_{ij} D_{ij} \bordermatrix{
                                                    ~ & & j & \cr
                                                      &0&x_j&0\cr} \\
                                        = & X (AB + (AB)^T) - X \circ (\widetilde{AC} + \widehat{AD}),
\end{aligned}
\end{equation}
where $\widetilde{AC}$ is defined in Eqn.~(\ref{equ:AC}) and $\widehat{AD}$ is defined in Eqn.~(\ref{equ:AD}).

\section{Gradients of View Specific SCNN}
\label{appdx:specific}

Although this paper doesn't use specific (asymmetric) SCNN for experiment, we give the gradients of cost function for reference. We denote the output of two sub-networks by $X=[\bfx_1, \bfx_2, \cdots, \bfx_n]$ and $Y=[\bfy_1, \bfy_2, \cdots, \bfx_m]$, where $n$ is the number of samples in view1 and $m$ is the number of samples in view2. The dimensions of their corresponding similarity matrix $S$ and mask matrix $M$ are $n \times m$.

Following the derivation process in Appendix~\ref{appdx:general}, the gradients of $J_{dev}$ with respect to $X$ and $Y$ are as follows.

\begin{equation}
\label{equ:bp-x1}
  \frac{\partial J_{dev}}{\partial X} = Y (EF)^T - X \circ \widetilde{EG},
\end{equation}
where
\begin{eqnarray}
E = -\alpha W \circ M \circ \frac{e^{-\alpha (S - \beta) \circ M}}{e^{-\alpha (S - \beta) \circ M} + 1}, \\
F = [F_{ij}]_{n \times n}, F_{ij}=\frac{1}{\sqrt{\bfx_i^T \bfx_i \bfy_j^T \bfy_j}},\\
G = [G_{ij}]_{n \times n}, G_{ij}=F_{ij} \frac{\bfx_i^T \bfy_j}{\bfx_i^T \bfx_i}, \\
\widetilde{EG} = repmat(\left(\sum_j E_{ij} G_{ij}\right)^T, d, 1). \\
\end{eqnarray}
And
\begin{equation}
\label{equ:bp-y1}
  \frac{\partial J_{dev}}{\partial Y} = X (EF) - Y \circ \widehat{EH},
\end{equation}
where
\begin{eqnarray}
H = [H_{ij}]_{n \times n}, H_{ij}=F_{ij} \frac{\bfx_i^T \bfy_j}{\bfy_j^T \bfy_j}, \\
\widehat{EH} = repmat(\sum_i E_{ij} H_{ij}, d, 1).
\end{eqnarray}

\section*{Acknowledgment}

This work was supported by the Chinese National Natural Science Foundation Projects \#61105023, \#61103156, \#61105037, \#61203267, \#61375037, National Science and Technology Support Program Project \#2013BAK02B01, Chinese Academy of Sciences Project No. KGZD-EW-102-2, Jiangsu Science and Technology Support Program Project \#BE2012627, and AuthenMetric R\&D Funds.

\ifCLASSOPTIONcaptionsoff
  \newpage
\fi

\bibliographystyle{IEEEtran}
\bibliography{vision}

\begin{IEEEbiography}[{\includegraphics[width=1in,height=1.25in,clip,keepaspectratio]{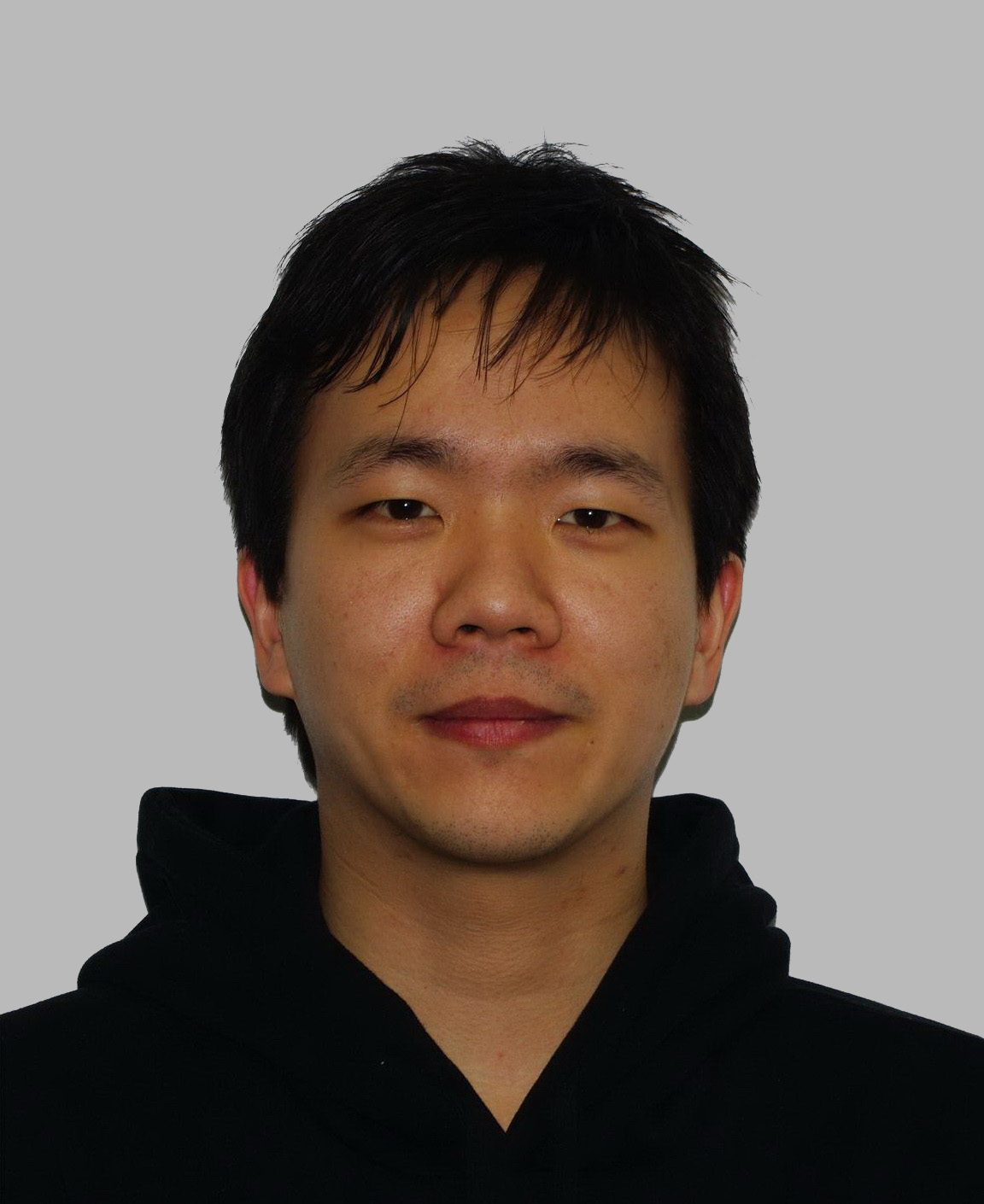}}]{Dong Yi}
received the B.S. degree in electronic engineering in 2003, the M.S. degree in communication and information system from Wuhan University, Wuhan, China, in 2006, and received the Ph.D. degree in pattern recognition and intelligent systems from CASIA, Beijing, China, in 2009. His research areas are unconstrained face recognition, heterogeneous face recognition, and deep learning. He has authored and acted as a reviewer for tens of articles in international conferences and journals. He developed the face biometric algorithms and systems for the Shenzhen-Hongkong immigration control project and 2008 Beijing Olympic Games.
\end{IEEEbiography}

\begin{IEEEbiography}{Zhen Lei}
received the B.S. degree in automation from the University of Science and Technology of China (USTC), in 2005, and the Ph.D. degree from the Institute of Automation, Chinese Academy of Sciences (CASIA), in 2010, where he is now an assistant professor. His research interests are in computer vision, pattern recognition, image processing, and face recognition in particular. He has published over 40 papers in international journals and conferences.
\end{IEEEbiography}


\begin{IEEEbiography}{Stan Z. Li}
received the B.Eng. degree from Hunan University, Changsha, China, the M.Eng. degree from the National University of Defense Technology, China, and the Ph.D. degree from Surrey University, Surrey, U.K. He is currently a Professor and
the Director of Center for Biometrics and Security Research (CBSR), Institute of Automation, Chinese Academy of Sciences (CASIA). He worked at Microsoft Research Asia as a researcher from 2000 to 2004. Prior to that, he was an Associate Professor at Nanyang Technological University, Singapore. His research interest includes pattern recognition and machine learning, image and vision processing, face recognition, biometrics, and intelligent video surveillance. He has published over 200 papers in international journals and conferences, and authored and edited eight books.
\end{IEEEbiography}

\end{document}